\documentclass[eat,twocolumn]{jmlr}

\usepackage{longtable}% for long tables

\usepackage{booktabs}
\usepackage[load-configurations=version-1]{siunitx} % newer version
\theorembodyfont{\upshape}
\theoremheaderfont{\scshape}
\theorempostheader{:}
\theoremsep{\newline}

\jmlrvolume{}
\firstpageno{1}
\jmlryear{2022}
\jmlrworkshop{Machine Learning for Health (ML4H) 2022}

\title[Self-training of Machine Learning Models for Liver Histopathology]{Self-training of Machine Learning Models for Liver Histopathology: Generalization under Clinical Shifts}

 \author{\Name{Jin Li,} \Name{Deepta Rajan,} \Name{Chintan Shah,} \Name{Dinkar Juyal,} \Name{Shreya Chakraborty,} \Name{Chandan Akiti,} \Name{Filip Kos,} \Name{Janani Iyer,} \Name{Anand Sampat,} \Name{Ali Behrooz}  \Email{ali.behrooz@pathai.com}\\
 \addr PathAI, Inc., Boston MA, USA
 }

\begin{document}

\maketitle

\begin{abstract}
Histopathology images are gigapixel-sized and include features and information at different resolutions. Collecting annotations in histopathology requires highly specialized pathologists, making it expensive and time-consuming. Self-training can alleviate annotation constraints by learning from both labeled and unlabeled data, reducing the amount of annotations required from pathologists. 
We study the design of teacher-student self-training systems for Non-alcoholic Steatohepatitis (NASH) using clinical histopathology datasets with limited annotations. We evaluate the models on in-distribution and out-of-distribution test data under clinical data shifts. We demonstrate that through self-training, the best student model statistically outperforms the teacher with a $3\%$ absolute difference on the macro F1 score. The best student model also approaches the performance of a fully supervised model trained with twice as many annotations.   

% \textcolor{red}{We explore four self-training approaches in the teacher-student paradigm including Noisy Student Training and Meta Pseudo Labels.} 

\end{abstract}

\section{Introduction}
\label{sec:intro}

A pathologist's diagnosis is often considered as the ground truth for disease detection, staging, and severity scoring. The traditional histopathology workflow involves manual reviewing of a large number of whole-slide images (WSI) to evaluate cell and tissue morphologies. The task is time-consuming and is complicated by high inter and intra-annotator variability \citep{javed2022rethinking}. Machine learning (ML) models have been shown to improve throughput of histopathology workflows in detection and prediction of cancer subtypes, gene mutations~\citep{coudray2018classification}, and molecular phenotypes~\citep{diao2021human}. However, the success of these approaches relies on the quantity and quality of annotations from board-certified pathologists, which are expensive to obtain. The large size of histopathology images also presents a challenge in getting exhaustive annotations. A successful way to navigate this problem is to extract patches from the WSI which has enabled the success of deep learning solutions \citep{srinidhi2020deep}. The intrinsic biases in acquiring histopathology images imply that sampling patches from WSIs rarely covers the full data distribution. There exists a number of known confounders causing batch effects~\citep{howard2021impact} that stem from variations across tissue sites, patient demographics, hospitals, stains, and scanners. There is also a severe label imbalance due to varying comorbidities within patient populations enrolled in different clinical trials \citep{thiagarajan2018understanding}.

Semi-supervised learning (SSL) for domain adaptation handles these data-driven constraints by leveraging unlabeled data through pseudo labeling \citep{lee2013pseudo} to develop accurate and robust models. Several prior works both in natural and medical imaging applications \citep{zou2018unsupervised,kamnitsas2021transductive,rajan2021self} show significant performance gains over fully-supervised methods. However, it is often difficult to exactly define under what conditions SSL works or which modeling components help or hurt performance \citep{van2020survey}. In addition, it is important to investigate how SSL methods can generalize to clinical shifts, as trained histopathology models need to do well on new clinical trials.

\section{Related Work}
\label{sec:related_work}

Prior work in semi-supervised learning for histopathology image classification has involved model pre-training on large-scale natural image datasets followed by fine-tuning as shown in \cite{marini2020semi,shaw2020teacher}, combining self-supervision with semi-supervision, or exploiting consistency regularization 
 in \cite{srinidhi2021self,pulido2020semi}. However, most of these techniques require designing domain-specific pre-training tasks and augmentations. 
 
%  Nearly all of these papers keep the teacher model frozen when training the student. \cite{kamnitsas2021transductive} demonstrate that improving the quality of selected unlabeled data for SSL can improve performance on MRI datasets. 

For AI-based analysis of NASH histopathology, most previous works deploy supervised models with convolutional neural networks (CNN) to segment disease features in WSIs from liver samples. A CNN-based WSI segmentation system was developed by \cite{taylor2021machine} to measure important histological characteristics of NASH, such as steatosis, inflammation, hepatocellular ballooning, and fibrosis.  \cite{natureNASHTrichrome} employed Masson's Trichrome WSIs to create 4 CNNs that categorize patches into stages of fibrosis, ballooning, inflammation, and steatosis. The logits from each model were averaged over all the patches of a trichrome WSI to produce the slide level classes. \cite{Dwivedi_2022_CVPR} showed improvements in AI-based NASH scoring by applying fusion models to segmentation masks from NASH Hematoxylin and Eosin (H\&E) and Trichrome samples. 

To the best of our knowledge, this work is the first effort to analyze the design of semi-supervised learning algorithms for evaluating samples from NASH patients using histopathology WSIs from clinical trials. We show that self-training improves performance over the supervised baseline on both the in-distribution dataset and out-of-distribution clinical trial datasets. We also demonstrate that by using unlabeled data, student models can approach the performance of fully supervised models trained with twice the number of labels.

% With the overarching goal of reducing annotation costs and maintaining robustness to clinical data shifts, we explore and apply \textcolor{red}{four different self-training modeling choices for histopathology. They include noisy student training (NST) by \cite{Xie2020SelfTrainingWN}, meta pseudo labels (MPL) by \cite{Pham2020MetaPL} and SSL methods proposed by \cite{french2017self} and \cite{marini2020semi}. The methods are outlined in Section $3$.} \textcolor{purple}{[remove]uncertainty-aware pseudo label selection (UPS) by \cite{Rizve2021InDO}}. 

\begin{figure*}[t]
  \centering
  \includegraphics[width=0.85\linewidth,height=\textheight,keepaspectratio]{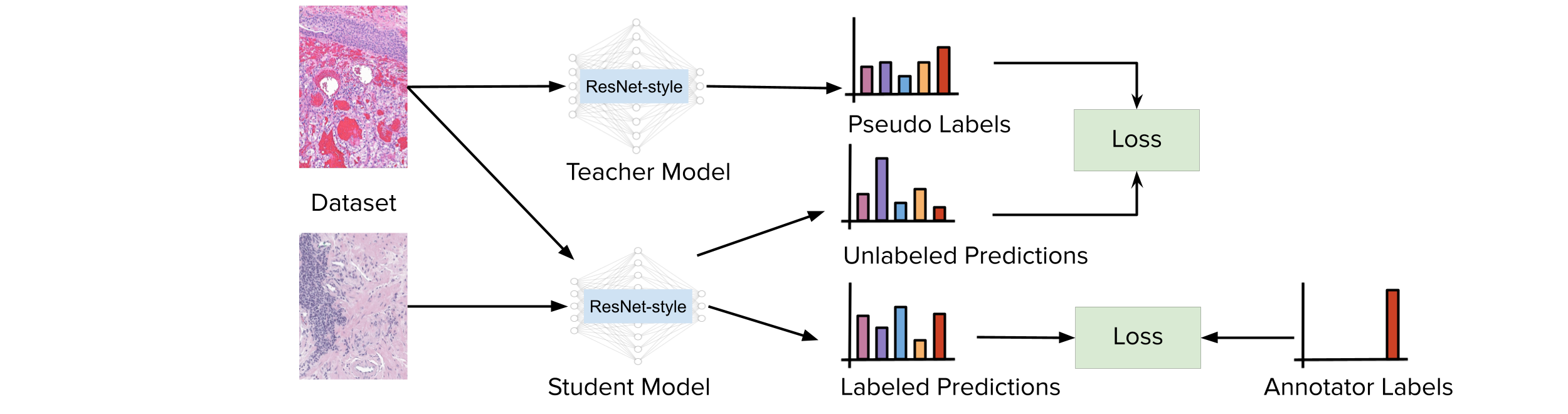}
  \vspace{0.1in}
  \caption{Diagram of how the student model is trained in the teacher-student paradigm. In the training step, the teacher model first creates pseudo labels from the unlabeled data. Then the student model is optimized from the sum of two losses: one based on the unlabeled predictions and the pseudo labels and the other from the labeled predictions and the annotator labels.}
  \label{fig:student_teacher_diagram}
\end{figure*}

\section{Methods}
Teacher-student learning paradigms leverage pseudo labels \citep{lee2013pseudo}, and involve the following sequence of steps: (i) train a teacher model on the labeled data, (ii) create pseudo labels on the unlabeled data, (iii) select pseudo labels by filtering them based on some criteria, and (iv) train a student model on both the labeled and pseudo labeled data (see Figure \ref{fig:student_teacher_diagram}). This training paradigm allows the teacher model to distill knowledge from the pre-trained teacher to the student \citep{Hinton2015DistillingTK}. We primarily focus on analyzing the NST and MPL approach and we use the other methods in Section 2 for comparison. See Appendix \ref{apd:teacher_student_methods} for a detailed description of the methods we used. See Appendix \ref{apd:modeling_framework} for the modeling framework, including the augmentations we used, calibration methods, model architecture, and pseudo label selection method. And see Appendix \ref{apd:training_hyperparameters} for the hyperparameters we picked to train our teacher student models.

\begin{table*}[th]
	\centering
	\renewcommand{\arraystretch}{1.1}
	\renewcommand{\tabcolsep}{5pt}
	\begin{tabular}{|l|c|c|c||l|c|c|c|}
		\hline
		\multicolumn{1}{|c|}{\textbf{Data}} &
		\textbf{Patients} & 
		\textbf{Slides} & 
		\textbf{Patches} \\ \hline
		Training  & 403 & 535  & 1,136,360  \\ 
		Validation & 44 & 60 & 60,256  \\  
		IID Test & 45 & 57 & 49,842  \\
		\hline
		CT A & 44 & 44 & 151,814  \\
		CT B & 14 & 14 & 41,688  \\
		CT C & 94 & 94 & 974,529 \\ \hline
% 		OOD Test & 44 & 44 & 25,975  & Disease & 0 & 0 & 0 \\ \hline
	\end{tabular}
	\caption{Summary of the histopathology image sample sizes used in our study. Clinical trial (CT) shifts A, B, C are out of distribution data.}
	\label{table:data-size}
\end{table*}

\section{Experiments}

\subsection{Dataset}
\noindent \textbf{Non-Alcoholic Steatohepatitis: } Non-Alcoholic Steatohepatitis (NASH) is a chronic liver disease linked to obesity and other metabolic diseases, leading to type II diabetes and fibrotic liver~\cite{williams2011prevalence}. Our dataset consists of digitized liver biopsy slides of varied sizes. These slides are stained by H\&E stain and scanned at 40x resolution by Aperio scanners. In this work, we consider the problem of classifying patches extracted from WSIs into one of $13$ classes associated with different histologic features in NASH. All slides are partially annotated by board certified pathologists trained in hepatobiliary pathology with one of the 13 classes. These 13 classes are: Bile Duct, Portal Inflammation, Steatosis, Normal Hepatocytes. Hepatocellular Swelling, Lobular Inflammation, Normal, Lumen, Blood Vessels, Interface Hepatitis, Hepatocellular Ballooning, Microvesicular Steatosis, Normal Interface. We sample patches of size $526$ x $526$ pixels at the highest resolution from these annotated regions for patch-based learning. The result of the segmentation can then be used in downstream analysis to predict the clinical scores or patient outcomes.

\noindent \textbf{Dataset Split:} Our empirical study comprises an in-distribution (ID) dataset and three generalization test sets. The WSIs belonging to the ID dataset are obtained from patients belonging to two different randomized control trials, and make up the source domain. We start with 201 patients in our train set, 44 patients in the validation set and 45 patients in our test set. We use patches from the slides of these patients to train the baseline teacher model. To train the student model, we use unlabeled patches from the slides of an additional 202 patients. To see how well our student model performs against a model trained on more labels,  we collect labels for those 202 patients and train another model (called the Oracle) on the patches from all 403 patients.

\noindent \textbf{Evaluation under Clinical Trial Shifts:} The intrinsic variations in data collection cause many challenging distribution shifts to co-exist in histopathology datasets. Potential causes can be shifts in patients population (age, ethnicity, gender, clinical history etc.), prevalence (stage of disease, target selection), data acquisition process (scanner, biopsy thickness, staining protocol) or in annotations (varying expertise of annotators or annotation policy). 

Our test set is divided into four subsets: in distribution test consists of images from the same clinical trials that were used in model training and validation. The other three test sets consist of images used in three different clinical trials (CT-A, CT-B, CT-C) that were not used during training. The sizes of the test sets are given in Table~\ref{table:data-size}. Based on NASH scores from human experts, the ID Test is the most severely diseased, while CT-B and CT-C are milder (earlier stage of disease); CT-A is mild in lobular inflammation (LI) and hepatocellular ballooning (HB) but severe in steatosis. There are significant differences in annotation distributions between these sets. While the ID Test has a fairly balanced distribution of classes, the distributions of CT-B and CT-C are very skewed towards non-pathological classes (Normal or Normal Hepatocytes). CT-A on the other hand is rich in annotations of LI and HB, as well as steatosis. Figure \ref{fig:class_freq} gives the distribution shift for the different classes across the various test datasets used.

\begin{table*}[htbp]
  \caption{Notation for models explored in this work. Note that we use (\texttt{Teacher}) to develop each of the students.}
  {%
    \begin{tabular}{|l|l|}
    \hline
    \textbf{Model} & \textbf{Description}\\ 
    \hline
    Teacher & Teacher with augmentation + mixup + dropout \\
    \hline
    SS+UL & Partial labeled + unlabeled loss (conditional entropy + class balance)  \\
    ST+FT & Pre-train student on pseudo labels and fine-tune on labeled data \\
    \hline
    NST & Iterative self-training with noisy students \\
    NST+T & NST with temperature scaling \\
    NST+T+U & NST with temperature scaling and UPS method \\
    MPL & Meta pseudo labels \\
    MPL+T & MPL with temperature scaling \\
    \hline
    Oracle & Trained using $100$\% labeled data + augmentation + mixup + dropout \\
    \hline
    \end{tabular}
  }
\label{table:notation}
\end{table*}
\setlength{\tabcolsep}{1.4pt}

\begin{table*}[t]
\centering
\caption{Performance of the teacher model, student models, and oracle across IDD Test and clinical trial shifts CT-A, CT-B, CT-C. Due to imbalance in the datasets, we consider the macro F1 score as the primary metric for deriving performance trends. The bounds reported are 95\% confidence intervals obtained by bootstrap analysis. Student models do much better than the teacher and MPL student can match the performance of the oracle on most clinical trial shifts.}
\renewcommand{\arraystretch}{1.5}
\renewcommand{\tabcolsep}{2.0pt}

\begin{tabular}{|c|c|c|c|c|c|c|c|c|}
\hline
% \multirow{2}{*}
{\textbf{Model}} &  
    \multicolumn{2}{c|}{\textbf{IDD Test}} & 
    \multicolumn{2}{c|}{\textbf{CT-A}} &
    \multicolumn{2}{c|}{\textbf{CT-B}} &
    \multicolumn{2}{c|}{\textbf{CT-C}} \\
    \cline{2-9}
% \hline
& \textbf{F1} & \textbf{Bounds} & \textbf{F1} & \textbf{Bounds} &  \textbf{F1} & \textbf{Bounds} &
\textbf{F1} & \textbf{Bounds}\\
\hline
Teacher & 0.492 & 0.487, 0.497 & 0.481 & 0.472, 0.491 & 0.328 & 0.318, 0.338 & 0.431 & 0.428, 0.435  \\
\hline
SS+UL & 0.482 & 0.478, 0.486 & 0.466 & 0.457, 0.475  & 0.309 & 0.300, 0.319 & 0.408 & 0.404, 0.411 \\
SS + FT & \textbf{0.525} & 0.520, 0.529 & 0.526 & 0.516, 0.536 & 0.327 & 0.316, 0.338 & 0.471 & 0.468, 0.474  \\
NST & 0.494 & 0.490, 0.498 & 0.540 & 0.529, 0.551 & 0.348 & 0.336, 0.360 & 0.460 & 0.456, 0.464  \\
NST+T & 0.496 & 0.492, 0.501 & 0.512 & 0.501, 0.523 & 0.340 & 0.325, 0.353 & 0.443 & 0.440, 0.446  \\
NST+T+U & 0.517 & 0.512, 0.520 & \textbf{0.549} & 0.538, 0.560 & 0.329 & 0.318, 0.341 & 0.467 & 0.463, 0.470  \\
MPL & 0.524 & 0.520, 0.529 & 0.525 & 0.516, 0.535 & \textbf{0.365} & 0.353, 0.377 & \textbf{0.474} & 0.470, 0.477  \\
MPL+T & 0.524 & 0.520, 0.529 & 0.526 & 0.516, 0.537 & 0.363 & 0.350, 0.376 & 0.467 & 0.463, 0.470  \\
\hline
Oracle & 0.522 & 0.518, 0.527 & 0.598 & 0.588, 0.607 & 0.367 & 0.356, 0.379 &  0.474 & 0.471, 0.478  \\
\hline

\end{tabular}
\label{table:st_metrics}
\end{table*}

\section{Results}
We first train the teacher models on patches from the slides of 201 patients and treat that as our baseline. We then used unlabeled patches from an additional 202 patients combined with the patches used in the teacher model to train all of our student models. Finally, we collect annotations for the 202 patients and use all 403 patients to train the oracle. We make the following observations based on our results:

\noindent \textbf{Student models outperform teacher models on both in distribution and out of distribution data} Based on Table \ref{table:st_metrics}, we find that the student models consistently perform better than the teacher model. For example, MPL is $3$\% better than Teacher on IDD Test set and around $4$\% better on CT-A, CT-B, and CT-C. This demonstrates that student models can effectively learn from unlabeled data and produce better results than the baseline teacher models.

\noindent \textbf{Self-training is nearly as good as collecting twice as many expert annotations} We compare one of our best student models, MPL, which only has labels from patches from 201 patients, to the oracle, which has labels from all 403 patients. Despite the fact that the oracle has access to twice as much labeled information as MPL, MPL is able to match the performance of the oracle on IDD (macro F1 score of $0.524$ vs $0.522$). Furthermore, the student model can match the performance on CT-B ($0.365$ vs $0.367$) and CT-C (both have macro F1 scores of 0.474). However, for CT-A, MPL was not able to match performance of the oracle ($0.526$ vs $0.598$). Even the best performing teacher-student training method (NST+T+U) on CT-A ($0.549$) is unable to match the oracle's performance on dataset CT-A.

\noindent \textbf{Evolving second generation students in NST yield greater benefits} In NST, we used the second generation student evolved from the first student, meaning we trained two student models and used the last one for inference. We find that evolving a second student generally adds 1-2\% improvement of macro F1 score on the test datasets but any further iterations did not improve performance.

\noindent \textbf{Calibrated methods do not necessarily improve student models} We tried to further improve our models by improving the calibration of our teacher models with temperature scaling and uncertainty-aware pseudo-label selection (UPS). A better calibrated model will produce better pseudo labels which can then improve which patches are selected during the pseudo label filtering process. The performance of calibrated students NST+T+U and MPL+T is not significantly different than the non-calibrated counterparts NST and MPL (Table \ref{table:st_metrics}). We suspect that even though the selected pseudo labeled samples may be better calibrated, they are not any more diverse and thus the student models will not benefit from those unlabeled patches.

\section{Conclusion}

Prohibitively expensive collection of annotations from domain experts and generalizability of trained models under clinical data shifts are two of the major challenges faced in histopathology image analysis and model development. To address this, we train different self-training models that leverages unlabeled data to improve performance on these generalization sets. We evaluate each method on data from three separate clinical trials not used in model training as well as on an in-domain test set. We showed that semi-supervised models can improve significantly over the supervised baseline both on in-domain and out-of-domain test sets. MPL+T model was the most consistent performer out of the methods we explored, beating the teacher on every test set and matching the performance of Oracle on all but one test set. 

In the future, we can explore how to better select unlabeled data to improve the performance of student models on generalization data sets. For example, we can try pre-training a SimCLR model only on unlabeled to get representations of the input images and then use those representations to select a diverse set of unlabeled patches to feed into the student model \cite{Chen2020ASF}. We can also try to add in unlabeled data from different clinical trials to see how much that improves generalization performance on out of distribution clinical trial shifts. 

% Furthermore, we can use methods in \cite{Stacke2021MeasuringDS} to measure the domain shift between data learned in the teacher model and unlabeled patches. And then we can select the unlabeled patches with the highest domain shift, train the student model with these patches, and then measure how well the student model performs on the generalization sets.

Overall, self-training has shown a potential to significantly reduce the amount of labeled data required in histopathology image analysis without compromising model performance and generalizability.

\acks{Thanks to Aryan Pedawi, Diksha, Tanmaya, John, Chaitanya, Zahil, and Aaditya
Prakash for their comments and support. Also, special thanks to Ben Trotter for
the excellent assistance on datasets, and to Murray Resnick and Fedaa Najdawi
for their expert inputs on histopathology.}

\bibliography{pmlr-sample}

\appendix

\section{Teacher Student Methods}\label{apd:teacher_student_methods}

\begin{figure*}[t]
  \centering
  \includegraphics[width=0.85\linewidth,height=\textheight,keepaspectratio]{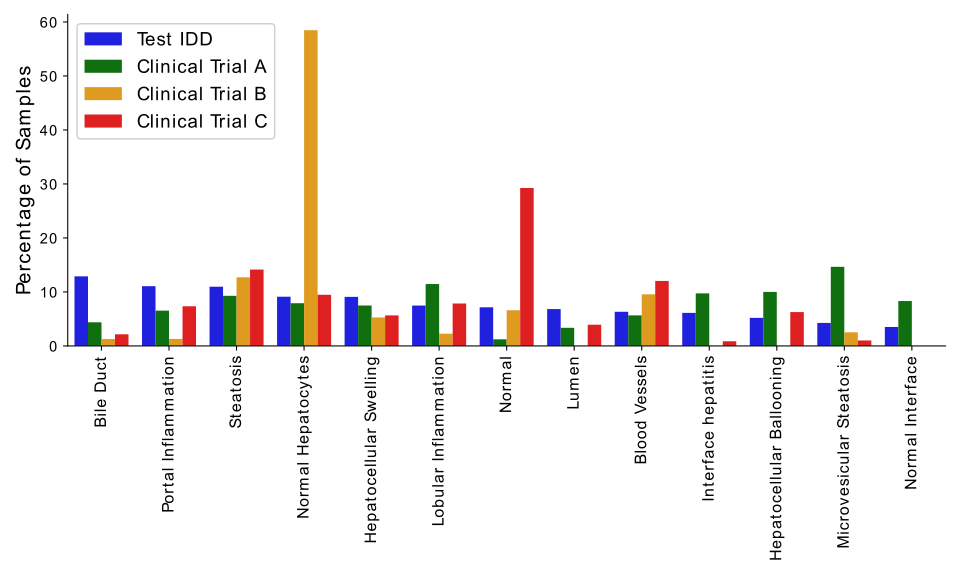}
  \vspace{0.1in}
  \caption{Percentage of samples belonging to each of the $13$ classes in the test sets.}
  \label{fig:class_freq}
\end{figure*}

Here we describe in detail the methods of teacher student training. Let $T$ and $S$ represent the teacher and student models with model weights $\theta_{T}$ and $\theta_{S}$ respectively. Let $D_{L} = \{ (x_{l}, y_{l}) \}_{i=1}^{N_{L}}$ be the labeled set with $N_{L}$ number of labeled samples, where $x_{i}$ denotes an input image patch and $y_{i} \in \mathbb{R}^{C} $ corresponds to the one-hot label with $C$ classes. Similarly, let $D_{U} = \{ (x_{u}) \}_{i=1}^{N_{U}} $ be the unlabeled set with a total of $N_{U}$ samples. Further, we denote $\hat{y} = T(x_{u}; \theta_{T} )$ as the soft predictions of the teacher model on unlabeled data (i.e. the pseudo-label) and define $S(x_{l}; \theta_{S})$ and $S(x_{u}; \theta_{S})$ as student predictions for the labeled and unlabeled data. Let $D_{P} = \{ (x_{u}, \hat{y}) \}_{i=1}^{N_{U}}$ be the pseudo labeled dataset. Lastly the cross-entropy between two distributions $a, b$ is represented by $CE(a, b)$. 

Formally, we first optimize the teacher weights as in equation \ref{eq:teacher_opt_l}, construct the pseudo labeled set $D_{P}$, and then filter pseudo labels to form a subset $D_{P^{*}}$. Finally, we train the student model on both the labeled and  pseudo labeled set $D_{B} = D_{L} \cup D_{P^{*}}$ as in equation \ref{eq:student_opt_l_ul}.

\begin{equation} 
\label{eq:teacher_opt_l}
  \theta_{T} = \underset{\rm \theta}{\rm argmin} \: \mathbb{E}_{x_{l}, y_{l}} [ CE( T(x_{l}; \theta); y_{l} ) ) ] 
\end{equation}

\begin{equation} 
\label{eq:student_opt_l_ul}
  \theta_{S} = \underset{\rm \theta}{\rm argmin} \: \mathbb{E}_{x_{i}, y_{i} \in D_{B}} [ CE( S(x_{i}; \theta); y_{i} ) ) ] 
\end{equation}

\subsection{Noisy Student Training (NST)}
NST proposed by \cite{Xie2020SelfTrainingWN} builds on the basic premise of teacher-student training by adding additional noise to the teacher or student models. The student models are also evolved over generations by turning the student into the teacher for subsequent iterations. We use both data and model noise, which includes data augmentation, dropout and mixup. We only perform two iterations of this training as students from subsequent generations did not noticeably improve performance similar to the finding in~\cite{rajan2021self}.

\subsection{Meta Pseudo Labels (MPL)}
The teacher's weights are not updated during student training in NST. \cite{Pham2020MetaPL} in their work trained the student and teacher simultaneously to allow the teacher to improve based on how well the student performs. In general, the optimal student parameters $\theta_{S}$ always depend on the teacher parameters $\theta_{T}$, since the student learns from the pseudo labels of the teacher. We refer to the original paper for more details on the algorithm.

\subsection{Semi-supervised Learning with Labeled + Unlabeled Loss (SS + UL)} 
\cite{french2017self} train their models with a cross-entropy loss for labeled samples and an additional loss for the unlabeled samples. In particular, a conditional entropy loss is used to force high-confidence predictions for unlabeled samples, and a class balance loss is used to deal with class imbalance. During our experiment, both unlabeled loss terms are given a weight of $0.2$. 

\subsection{Semi-supervised Learning with Fine-tuning (SS + FT)} 
We also compare the previous methods to another SSL approach proposed by \cite{marini2020semi} in the context of histopathology. The teacher is trained from scratch on the labeled data and used to produce pseudo labels for the unlabeled data. However, in this case, the student is pre-trained on the filtered pseudo-labels and further fine-tuned on the labeled set.

\section{Modeling Framework}\label{apd:modeling_framework}
We provide descriptions of various factors that impact performance of these self-training algorithms in the following sections and demonstrate how well self-training methods can generalize to unseen distributions. 

\noindent \textbf{Augmentations:} During training, we augment both labeled and unlabeled patches for all models using flip, rotation, color distortions(brightness, contrast, saturation, hue) and noise corruptions. No augmentations were performed during test. 

\noindent \textbf{Calibration:} We use two calibration techniques to ensure the trained models are more reliable under data shifts. (i) Mixup \citep{Zhang2018mixupBE} is a popular approach that linearly interpolates between a pair of samples and their corresponding labels. Mixup enables regularization of neural network and improves model calibration as shown in \cite{thulasidasan2019mixup}. (ii) Temperature Scaling \citep{Hinton2015DistillingTK} scales the logits output from a network by a single parameter to adjust the entropy of a prediction.

\noindent \textbf{Model Architecture:} The backbone architecture for all models used in our study is a custom $21$-layer, ResNet-style convolutional neural network with residual connections. Each block consists of a convolution layer, a batch norm layer, and a ReLU activation unit. The model has about $13.2$ million parameters. Motivated by~\cite{oliver2018realistic}, we use the same network for both the teacher and the student to evaluate benefits of self-training without introducing confounding factors like model capacity. We detailed more of the model hyperparameter choices in Appendix \ref{apd:training_hyperparameters}.

\noindent \textbf{Pseudo-label Selection:} A crucial step in  teacher-student self-training is choosing the criteria for filtering pseudo labels to reduce the noise from the unlabeled data. We consider two different approaches to select pseudo labels. (i) Confidence thresholding : We remove pseudo labels if their confidence (probability of the most likely class) is less than some threshold $k$; (ii) Uncertainty-aware Pseudo-label Selection (UPS) : We remove pseudo labels where their MC dropout uncertainty is above a threshold \citep{Rizve2021InDO}. To compute the uncertainty for a single sample, we perform multiple forward passes with dropout enabled to produce a set of predictions and then compute the standard deviation of those predictions \citep{Gal2016DropoutAA}. 

\section{Training hyperparameters}\label{apd:training_hyperparameters}

We tune the teacher model for optimal hyperparameters based on the best macro F1 on the validation set and then use the same hyperparameters for the student models. Any additional hyperparameters  specific to the student model, such as temperature and confidence threshold, are tuned on the same validation set. For the teacher model, we conducted ablation studies using augmentations, calibration techniques and dropout. The teacher model that contained augmentations, mixup and dropout performed best, and this model was subsequently used for training students.

All models are trained using the Adam optimizer with a learning rate of $0.0001$, a cross entropy loss objective, and a dropout of $0.5$. The momentum for batch norm is set to $0.6$. We also set the maximum number of iterations to $30,000$ and decay the learning rate by $0.5$ every $10,000$ steps. We use a batch size of $128$ for teacher models, and student models use a batch size of $64$ for both labeled and unlabeled data. When using mixup, the amount of interpolation controlled by the $\alpha$ parameter is set to $0.2$ \citep{Zhang2018mixupBE}. For NST, we used a confidence threshold of $0.4$, a temperature of $1.05$, and evolved at most $2$ generations of student models. For MPL, we used a confidence threshold of $0.2$ and a temperature of $1.10$. For all teacher-student training methods, we use soft pseudo labels instead of hard pseudo labels since soft pseudo labels performed better in our experiments. When using the UPS method, the uncertainty threshold is set to $0.10$ and we performed $10$ forward passes for computing the standard deviation.

\setlength{\tabcolsep}{4pt}

\end{document}